\pdfoutput=1
\documentclass[11pt]{article}

\usepackage[preprint]{coling}

\usepackage{times}
\usepackage{latexsym}
\usepackage{hyperref}
\usepackage[T1]{fontenc}
\usepackage[utf8]{inputenc}

\usepackage{microtype}

\usepackage{inconsolata}

\usepackage{graphicx}
\usepackage{amssymb}
\usepackage{amsfonts}
\usepackage{amsmath}
\usepackage{booktabs}
\usepackage{colortbl} % 引入colortbl包
\usepackage{natbib}   % 引入natbib包，支持\citep命令

\usepackage{arydshln} % For \hdashline
\usepackage{makecell} % For \Xhline
\usepackage{multirow} 
\usepackage{graphicx}
\usepackage{subcaption} % Include this package
\title{ReLayout: Towards Real-World Document Understanding via \\
Layout-enhanced Pre-training}

\author{Zhouqiang Jiang\(^1\), Bowen Wang\(^2\)\thanks{Corresponding author.}, Junhao Chen\(^1\), Yuta Nakashima\(^2\) \\[1ex] 
        Osaka University, Japan  \\[1ex] 
        \(^1\)\texttt{\{zhouqiang, junhao\}@is.ids.osaka-u.ac.jp} \\
        \(^2\)\texttt{\{wang, n-yuta\}@ids.osaka-u.ac.jp}}

\begin{document}
\maketitle
\begin{abstract}
Recent approaches for visually-rich document understanding (VrDU) uses manually annotated semantic groups, where a semantic group encompasses all semantically relevant but not obviously grouped words. As OCR tools are unable to automatically identify such grouping, we argue that current VrDU approaches are unrealistic. We thus introduce a new variant of the VrDU task, real-world visually-rich document understanding (ReVrDU), that does not allow for using manually annotated semantic groups. We also propose a new method, ReLayout, compliant with the ReVrDU scenario, which learns to capture semantic grouping through arranging words and bringing the representations of words that belong to the potential same semantic group closer together. Our experimental results demonstrate the performance of existing methods is deteriorated with the ReVrDU task, while ReLayout shows superiour performance. 

%Starting from the perspective of comprehensively enhancing layout information in representation, we introduce two layout-enhanced pre-training tasks: 1D Local Order Prediction (1-LOP) and 2D Text Segment Clustering (2-TSC). The 1-LOP masks global 1D position input and reconstructs local 1D position. The 2-TSC employs advanced self-supervised technique to adaptively bring closer the representations of text segments belonging to the same complete semantic group. These two pre-training methods learn comprehensively layout-enhanced multimodal representation. Moreover, ReLayout demonstrates strong robustness based on different layout inputs extracted by various real-world OCR tools. Independent of any manually-annotated layout input and solely using the simplest global 1D positions and word-wise 2D bounding boxes, ReLayout achieves leading performance in downstream tasks such as form and receipt understanding and document visual question answering.
\end{abstract}

\section{Introduction}
Modern visually-rich document understanding (VrDU), which aims at automating information extraction from visually-rich documents, has become an important research direction \citep{liu2019graph, jaume2019funsd, xu2020layoutlm, garncarek2021lambert, gu2022xylayoutlm, tu2023layoutmask}. Visually-rich documents, such as invoices, receipts, reports, and academic papers, not only contain a substantial volume of text data but also encode essential semantics needed for understanding them into their structures or \textit{layout}. Figure \ref{fig:example}(top) shows an example of such a document. People can perhaps group up ``Case type'' and ``Plaintiff's counsel'' into respective semantic groups, and also associate ``Case type'' and ``Asbestos'' as well as ``Plaintiff's counsel'' and the name and address on its right, even though they are spatially apart from each other, because of the knowledge on the layout (e.g., a field label and its content are typically placed next to each other). Such structural comprehension of documents provides strong prior on the semantics that respective layout groups have, enhancing the accuracy of information extraction.

\definecolor{BQ}{HTML}{F08275}
\definecolor{BA}{HTML}{96C279}
\begin{figure}[t]
    \centering
    \centering
    \includegraphics[width=\columnwidth]{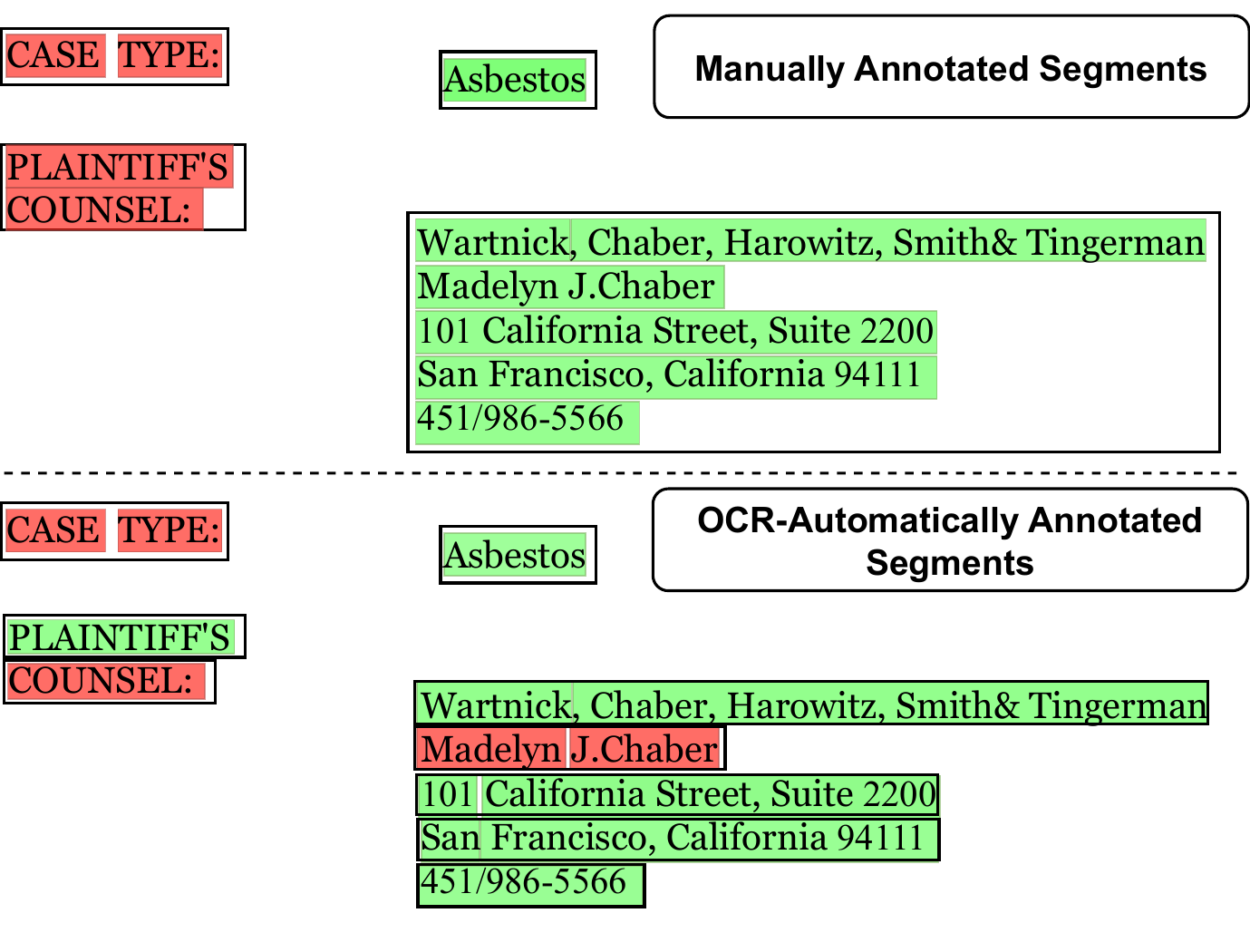}
    \caption{Result of LayoutMASK \cite{tu2023layoutmask} for Semantic Entity Classification using different segments. \framebox[0.4cm][c]{\rule{0pt}{0.15cm}}:Segment, \colorbox{BQ}{\phantom{0}}:\textbf{QUESTION},  \colorbox{BA}{\phantom{0}}:\textbf{ANSWER}.  }
    \label{fig:example}
\end{figure}

Early work \citep{hwang2019post, denk2019bertgrid} makes use of the empirical knowledge about layout that text flows from left to right and top to bottom. Such flow can be represented by global 1D positions associated with all words in OCR text. The text flow can be more precise when the document structure is obtained from auxiliary sources, such as XMLs' or PDFs' metadata \cite{wang2021layoutreader}. Recent studies have explored the interaction between OCR text and its layout in document images through pre-training model \citep{xu2020layoutlm, li2021structurallm, hong2022bros, tu2023layoutmask}. LayoutLM \citep{xu2020layoutlm} is the first to introduce word-wise 2D bounding boxes as layout embedding in the pre-training model. Similarly to BERT, LayoutLM masks words extracted by OCR but retains the corresponding layout embeddings, requiring the model to reconstruct the original words. The representations learned in this way exhibit excellent fine-tuning performance.

Subsequent pre-training approaches have shown that by incorporating the semantically relevant text blocks (semantic groups) and using a common segment (segment-wise) 2D bounding box as a layout embedding \cite{li2021structurallm}, richer semantic concepts can be provided, as illustrated in Figure \ref{fig:example}(top), thereby significantly enhancing performance. LayoutMASK\cite{tu2023layoutmask} further uses segment-wise 2D bounding boxes to change the global 1D position to an loacl 1D position within segments, enhancing the local information of text flow and further improving performance.

In the context of document comprehension, use of the structural information in documents faces an inherent challenge: \textit{semantic groups can facilitate automated document understanding, while they are manually annotated, yet we aim to achieve automated document understanding.}, which forms a paradox that the costly annotation of semantic groups and automated document understanding cannot coexist. Prior works \cite{li2021structurallm, huang2022layoutlmv3,tu2023layoutmask} avoid this problem using human-annotated semantic groups during fine-tuning (Figure \ref{fig:example}); however, semantic groups as accurate as human annotations are not available in real-world scenarios. Off-the-shelf OCR can group some spatially consecutive words (referred to as a \textit{text segment}) in a document as in Figure \ref{fig:example}(bottom), but they do not necessarily align with the actual semantics that the document encompasses.

To nourish exploration toward real-world VrDU, we propose a new VrDU task, coined \textit{ReVrDU} (\underline{\textbf{Re}}al-world \underline{\textbf{VrDU}}), on top of existing ones, which only allows for using information available from off-the-shelf OCR tools for both pre-training and fine-tuning, i.e., words, global 1D positions, word-wise 2D bounding boxes, and text segments, so that VrDU can be evaluated in alignment with real-world scenarios. 

We also propose a new pre-training model for ReVrDU, referred to as \textit{ReLayout} (\underline{\textbf{Re}}al-world \underline{\textbf{Layout}}-enhanced pre-training), use simple global 1D positions and word-wise 2D bounding boxes as layout input. In addition to the masked language modeling (MLM) strategy, ReLayout adopts 1D Local Order Prediction (1-LOP) and 2D Text Segment Clustering (2-TSC) strategies. The former reconstructs word order within each text segment. Through this task, a model learns local information about text flow as well as relationships cross text segments as it needs to predict where a text segments starts and ends in the masked global 1D positions. With the latter, a model learns to complete potential semantic groups information from text segments in a self-supervised manner. We experimentally show that pre-training a model with ReLayout demonstrates excellent downstream performance in both ideal and real-world scenarios.

% \textbf{Contribution}. We argue that research efforts so far toward layout-enhenced VrDU adopts a less realistic assumption that semantic groups of recognized text are available. ReVrDU offers a task definition, on top of existing datasets, for rigorously evaluating a model's performance in a realistic scenario. Our ReLayout pre-training strategy mitigates the challenges due to missing semantics grouping by letting a model learn local text flow information, relationships cross text segments and grouping semantically close text segments. \textcolor{red}{Some insights based on the experiments.}

\section{Related Work}
\subsection{Multimodal Pre-training}
Multimodal self-supervised pre-training models \cite{hong2022bros, wang2022lilt, xu2020layoutlm, xu2020layoutlmv2, powalski2021going, appalaraju2021docformer, xu2021layoutxlm, li2021structext, lee2022formnet, huang2022layoutlmv3, peng2022ernie}, due to their successful application across document layout, text, and visual modalities, have propelled rapid advancements in the field of VrDU. LayoutLM \cite{xu2020layoutlm} first introduces each token's 2D bounding box as layout embedding to enhance MLM. On top of LayoutLM, BROS \cite{hong2022bros} proposes a more challenging MLM task that masks larger regions. StructurelLM \cite{li2021structurallm} pre-trains a model by predicting positions of equally-sized regions in a document.  These pre-training tasks jointly model the relationships between text and the layout in documents. 

Given the richness of visual cues in image attributes such as fonts, colors, logos, and dividing lines in tables, many works incorporate the visual modality pre-training tasks \cite{gu2021unidoc, li2021selfdoc, xu2020layoutlmv2, huang2022layoutlmv3, gu2023unified}, including masked visual-language modeling \cite{xu2020layoutlmv2}, masked image modeling \cite{huang2022layoutlmv3}, word-patch alignment \cite{huang2022layoutlmv3}, text-image alignment \cite{xu2020layoutlmv2}, text-image match \cite{xu2020layoutlmv2}, and visual contrastive learning \cite{gu2023unified}. These tasks exploit the knowledge in visual components, providing additional performance boosts in the VrDU tasks.

ReLayout, however, exclusively uses text and layout modalities to evaluate the performance of models in ideal and real-world scenarios.

\subsection{Layout Information}
How to handle layout information is crucial for VrDU. LayoutLM first introduces spatial layout information into VrDU using word-level 2D bounding boxes. BROS proposes to encode relative spatial relationships of word-wise 2D bounding boxes with a  BERT-based model. StructureLM utilizes segment-wise 2D bounding boxes rather than word-wise to represent layout, showing promising performance improvements. LayoutMASK \cite{tu2023layoutmask} introduces a strong prior about layout through local 1D positions and segment-wise 2D bounding boxes. It uses segment-wise 2D bounding boxes to identify semantic groups and local 1D positions to guide the model in scanning tokens in the correct order. This method achieved SOTA performance on downstream tasks with manually annotated semantic groups, but obtaining semantic groups in real-world scenarios is impractical. 

ReLayout does not use complete semantic groups, which is practically unavailable, but supplies text segments that are automatically obtained with common OCR tools and do not necessarily align with actual semantic groups during pre-training. Pre-training strategy in ReLayout is designed to handle such text segments by learning to merge semantically similar text segments.

\begin{figure*}[t]
    \centering
    \includegraphics[width=\textwidth]{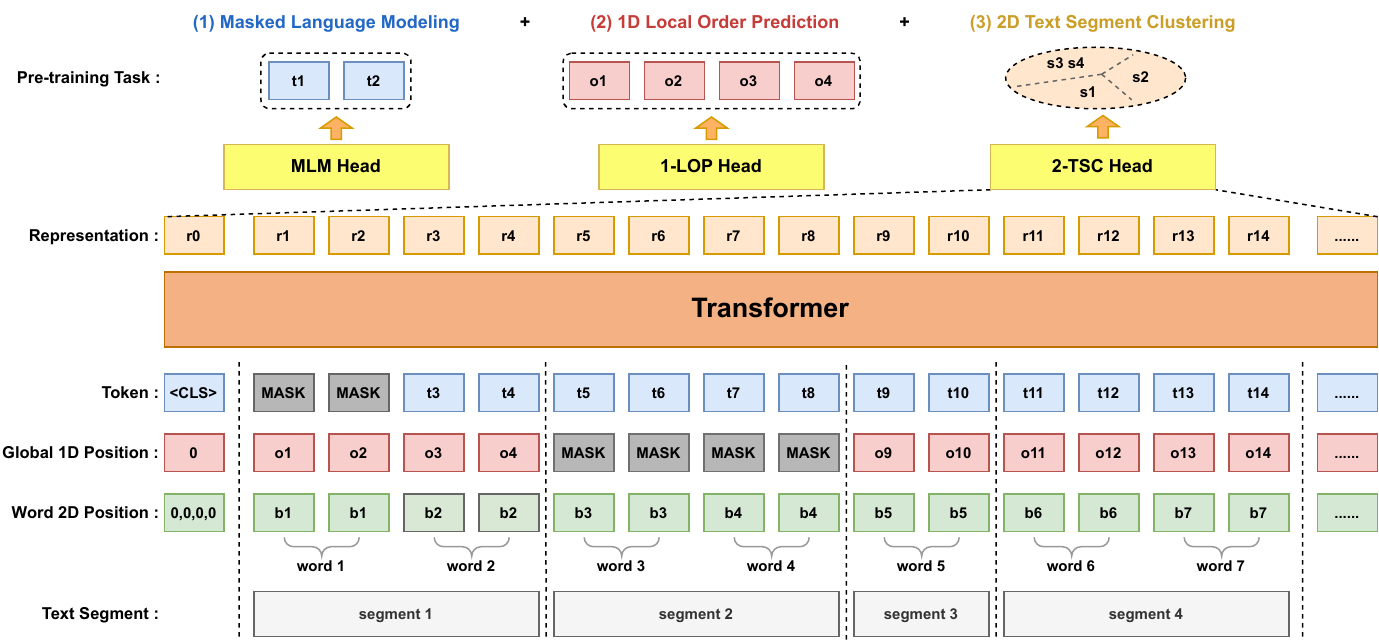}
    \caption{Architecture of ReLayout: MLM masks word-level tokens and reconstructs the original tokens. 1-LOP masks global 1D positions at the text segment and reconstructs local 1D positions. 2-TSC uses self-supervised techniques to adaptively cluster the representations of text segments that belong to the same semantic group.}
    \label{fig:arch}
\end{figure*}

\section{Task Definition: ReVrDU}

Our ReVrDU task is built on top of existing VrDU tasks that supply human-annotated semantic groups as input \cite{jaume2019funsd, park2019cord}. 
% To comply with a real-world fine-tuning scenario, in which OCR tools cannot provide perfect semantic grouping, we remove the reliance on the semantic groups.

Formally, traditional VrDU tasks typically provide a set $\mathcal{W} = \{w_l\}_{l = 1}^L$ of words, a set $\mathcal{O} = \{o_l\}_{l=1}^L$ of the corresponding global 1D positions $o_l \in \mathbb{Z}_{\geq 0}$, where $\mathbb{Z}_{\geq 0}$ is the set of non-negative integers, semantic groups $\mathcal{S}' = \{s'_k\}_{k = 1}^K$, where $s_k$ is the $k$-th semantic group that contains all words in the group (i.e., $s'_k \subset \mathcal{W}$), and a set $\mathcal{B} = \{b_l\}_{l = 1}^L$ of word-wise bounding boxes. Word $w_l$ is associated with word-wise bounding box $b_l \in \mathbb{R}^4$, represented by its top-left and bottom-right corners' positions. The semantic groups serves as a strong cue for understanding the semantics.

ReVrDU provides data in the same format, but to align with real-world scenarios, all data come from OCR results, e.g., a semantic group is replaced with just a set of consecutive words in a line (i.e., a text segment), which can be obtained from OCR tools. We denote a set of text segments by $\mathcal{S} = \{s_k\}_{k=1}^K$, where $s_k \subset \mathcal{W}$ is the $k$-th text segment.

\section{Method: ReLayout}
The unavailability of accurate semantic grouping in ReVrDU hinders the naive application of existing methods tailored for VrDU. Our ReLayout, a layout-enhanced multimodal pre-training model, effectively incorporates structural information in documents by pre-training a model to strengthen the understanding of local layout structures and relationships and learn semantic grouping via our proposed 1-LOP and 2-TSC strategies.

As shown in Figure~\ref{fig:arch}, we use a vanilla Transformer encoder \cite{vaswani2017attention} architecture as the backbone of our model.

\subsection{Tokenizers}

We use byte-pair encoding \cite{sennrich2015neural} to tokenize $\mathcal{W}$ into a set $\mathcal{T} = \{t_n\}_{n=1}^N$ of tokens $t_n$. We also reassign the global 1D positions according to $\mathcal{O}$ and the tokenization result. We denote the reassigned set of the global 1D positions as $\mathcal{O}' = \{o'_n\}_{n=1}^N$, where $o'_n \in \mathbb{Z}_{\geq 0}$. We also remap the bounding boxes $\{b_l\}$ to $\{b'_l\}$ and the text segments so that tokens derived from the same word have the same bounding box and are included in the same text segment, as shown in Figure \ref{fig:arch}.

\subsection{Embeddings} 

For tokens, we use a token embedding layer, denoted by $e^\text{t}_n = \text{TE}(t_n)$. The global 1D positions, represented by non-negative integers, are encoded with 1D position embedding layer, denoted by $e^\text{o}_n = \text{PE1D}(o'_n)$. The bounding box $b'_n$ is represented by 2D position embedding $e^\text{b}_n = \text{PE2D}(b'_n)$
The $n$-th input embedding $e_n$ to the backbone network is the sum of these embeddings, i.e.,
\begin{align}
    e_n = e^\text{t}_n + e^\text{o}_n + e^\text{b}_n.
\end{align}

\subsection{Pre-training Tasks}
We employ three pre-training tasks, i.e., MLM, 1-LOP, and 2-TSC pre-training tasks.

\subsubsection{Masked Language Modeling}
MLM is utilized to enable the model to learn multimodal representations of text-layout interactions by combining text and layout cues. We randomly mask tokens at the word level with given probability $P_\text{MLM}$, where tokens to be masked are replaced with \texttt{[mask]} token.

The all embeddings $e$ are then fed into the model, and the output representations pass through a non-linear MLM Head layer, obtaining logits for each masked token. With softmax, we compute the reconstructed probability \(p_j\) of the $j$-th masked tokens over the vocabulary ($j=1,\dots,J$). The loss function is thus defined as:

\begin{equation}
L_\text{MLM}=-\frac{1}{J}\sum_{j=1}^J{\log p_{j}}
\end{equation}

\subsubsection{1D Local Order Prediction}
Local 1D positions in each semantic group give a model some ideas about how the words are arranged in the document. As semantic grouping is not available in ReVrDU, we instead predict them within each text segment through the 1-LOP pre-training task. This design choice is not optimal as text segments do not necessarily correspond to semantic groups; we still consider that they can serve as a good proxy of semantic groups for learning local structure.\footnote{We experimentally validate this assumption.} By learning when to increment the position value and reset it to 1, our model grasps both within- and cross-segment local structures.
%however, to better understand the local structure of documents, local 1D positions of words in a semantic group are beneficial \cite{tu2023layoutmask}. Unfortunately, obtaining semantic groups in large-scale unlabeled datasets is impractical. 

As shown in Figure~\ref{fig:arch}, we randomly select some text segments with probability $P_\text{1-LOP}$ and mask all global 1D positions within all selected text segments. The output representations from the model then go through a non-linear 1-LOP head to predict the local 1D positions in the text segments. Letting $M$ be the number of masked tokens in total and $q_m$ the probability of the correct local position, the loss is defined as:
\begin{equation}
L_\text{1-LOP}=-\frac{1}{M}\sum_{m=1}^M{\log q_{m} }.    
\end{equation}
%where $q_{m}^{'}$ represents the predicted probability corresponding to the $m$-th local 1D order in this text segment. The overall $L_\text{1-LOP}$ will be averaged across all segments.

\subsubsection{2D Text Segment Clustering}
The 1-LOP pre-training task uses less accurate text segments provided by an OCR tool. A model pre-trained with such text segments may not be fully consistent with actual semantic groups, resulting in fragmented representations, as shown in Figure \ref{fig:example}(top). We thus wish a model to be more aware of semantic grouping. For this, we make the mild assumption that a semantic group consists of words (or text segments) that are semantically relevant and are spatially close to each other in the document. Under this assumption, we propose the 2-TSC pre-training task to help the model learn semantic grouping.

We borrow the idea from SimSiam \cite{chen2021exploring}, a type of contrastive learning that do not require negative samples, to let \textit{text segment representations} belonging to the potential same semantic group close to each other, where a text segment representation is the average pooling of the token representations in a text segment.

Let $\mathcal{R}_k = \{r_{ki}\}_{i=1}^{I}$ denote the set of representation vectors for the $i$-th token in the $k$-th text segment $s_k$. We represent the semantics of $s_k$ by average-pooling its token representations, i.e.:
\begin{equation}
v_k=\frac{1}{|\mathcal{R}_k|} \sum_{i=1}^{I} r_{ki}.
\end{equation}
We can find a set $K = \{(k, k')\}$ of semantically close text segment indices $k$ and $k'$, which satisfies the following two conditions with predefined thresholds $\theta_\text{dis}$ and $\theta_\text{sim}$:
\begin{align}
    \text{Dist}(s_k, s_{k'}) < \theta_\text{dis},\;\text{Sim}(v_k, v_{k'}) > \theta_\text{sim}.
\end{align}
$\text{Dist}(s_k, s_{k'})$ gives the Euclid distance between the centers of bounding boxes that encompasses all tokens in $s_k$ and $s_{k'}$, which can be computed based on the merged bounding box in the k-th text segment. $\text{Sim}$ gives the cosine similarity.
Given $(k, k') \in \mathcal{K}$, $v_k$ and $v_{k'}$ should be close to each other. We thus introduce a predictor \cite{chen2021exploring} to map $v_k$ to the same dimensional space as
\begin{equation}
z_k=\frac{1}{|\mathcal{R}_k|} \sum_{i=1}^{I} f(r_{ki}),
\end{equation}
and bring them closer by
\begin{equation}
L_\text{2-TSC} = -\text{Sim}( z_k, \text{stopgrad}( v_{k'})),
\end{equation}

With the above three pre-training objectives, the model is pre-trained with the following loss:
\begin{equation}
    L_\text{total}=L_\text{MLM}+\alpha L_\text{1-LOP}+\gamma L_\text{2-TSC}
\end{equation}
where $\alpha$ and $\gamma$ are hyper-parameters. $L_\text{2-TSC}$ is used only in the final epoch of pre-training.

\definecolor{BF}{HTML}{D9D9D9}
\begin{table*}[t]
\small
\centering
\begin{tabular}{lclccc}
\toprule
\multicolumn{1}{c}{\textbf{Method}}   & \textbf{\#Parameters} & \textbf{Modality} & \textbf{FUNSD(F1$\uparrow$)} & \textbf{CORD(F1$\uparrow$)} & \textbf{DocVQA(ANLS$\uparrow$)} \\ \midrule
BERT$_\mathrm{Base}$   \citep{devlin2018bert}           & 110M                  & T                 & 60.26          & 89.68         & 63.72          \\
RoBERTa$_\mathrm{Base}$ \citep{liu2019roberta}           & 125M                  & T                 & 66.48          & 93.54         & 66.42              \\
UniLMv2$_\mathrm{Base}$    \citep{bao2020unilmv2}       & 125M                  & T                 & 68.90          & 90.92         & 71.34          \\
LayoutLM$_\mathrm{Base}$   \citep{xu2020layoutlm}        & 160M                  & T+L+I             & 79.27          & -             & 69.79         \\
LayoutLMv2$_\mathrm{Base}$ \citep{xu2020layoutlmv2}       & 200M                  & T+L+I             & 82.76          & 94.95         &  \textbf{78.08}         \\
DocFormer$_\mathrm{Base}$ \citep{appalaraju2021docformer}        & 183M                  & T+L+I             & 83.34          & 96.33         & -              \\
BROS$_\mathrm{Base}$   \citep{hong2022bros}             & 110M                  & T+L               & 83.05          & 95.73         & 71.92          \\
\rowcolor{gray!30} LiLT$_\mathrm{Base}$    \citep{wang2022lilt}          & -                     & T+L               & 88.41          & 96.07         & -              \\
LayoutLMv3$_\mathrm{Base}$  \citep{huang2022layoutlmv3}       & 133M                  & T+L+I             & 81.61\textsuperscript{†}          & 94.64\textsuperscript{†}         & 74.56\textsuperscript{†}              \\ 
\rowcolor{gray!30} LayoutLMv3$_\mathrm{Base}$  \citep{huang2022layoutlmv3}       & 133M                  & T+L+I             & 90.29          & 96.56         & 78.76              \\ 
LayoutMask$_\mathrm{Base}$ \citep{tu2023layoutmask}  & 182M                   & T+L               & 73.97\textsuperscript{†}    & 82.37\textsuperscript{†}    & 70.79\textsuperscript{†}     \\ 
\rowcolor{gray!30} LayoutMask$_\mathrm{Base}$ \citep{tu2023layoutmask}  & 182M                   & T+L               & 92.91     & 96.99    &  - \\ \hline
\vspace{2pt}
\textbf{ReLayout$_\mathrm{Base}$ (Ours)}  \rule{0pt}{9pt}      & 125M              & T+L       & \textbf{84.64}    & \textbf{96.82}    &   76.02  \\ \hline

BERT$_\mathrm{Large}$    \citep{devlin2018bert}               & 340M                  & T                 & 65.63          & 90.25         & 67.45          \\
RoBERTa$_\mathrm{Large}$ \citep{liu2019roberta}             & 355M                  & T                 & 70.72          & 93.80         & 69.52               \\
UniLMv2$_\mathrm{Large}$     \citep{bao2020unilmv2}          & 355M                  & T                 & 72.57          & 92.05         & 77.09          \\
LayoutLM$_\mathrm{Large}$  \citep{xu2020layoutlm}    & 343M                  & T+L               & 77.89          &-               & 72.59          \\
LayoutLMv2$_\mathrm{Large}$  \citep{xu2020layoutlmv2}        & 426M                  & T+L+I             & 84.20           & 96.01         &  \textbf{83.48}          \\
DocFormer$_\mathrm{Large}$  \citep{appalaraju2021docformer}        & 536M                  & T+L+I             & 84.55          & 96.99         & -                \\
BROS$_\mathrm{Large}$ \citep{hong2022bros}              & 340M                  & T+L               & 84.52          & 97.40          & 74.70                \\
\rowcolor{gray!30} StructuralLM$_\mathrm{Large}$ \citep{li2021structurallm}              & 355M                  & T+L               & 85.14          & -         & 3.94$^\ddagger$                \\
LayoutLMv3$_\mathrm{Large}$  \citep{huang2022layoutlmv3}       & 368M                  & T+L+I             & 84.13\textsuperscript{†}          & 96.88\textsuperscript{†}         & 78.26\textsuperscript{†}               \\
\rowcolor{gray!30} LayoutLMv3$_\mathrm{Large}$  \citep{huang2022layoutlmv3}       & 368M                  & T+L+I             & 92.08          & 97.46         & 83.37               \\
LayoutMask$_\mathrm{Large}$ \citep{tu2023layoutmask}  & 404M                   & T+L          & 78.12\textsuperscript{†}     & 84.67\textsuperscript{†}    & 74.06\textsuperscript{†}  \\ 
\rowcolor{gray!30} LayoutMask$_\mathrm{Large}$ \citep{tu2023layoutmask}  & 404M                   & T+L          & 93.20     & 97.19    & -  \\ \hline
\vspace{2pt}
\textbf{ReLayout$_\mathrm{Large}$ (Ours)} \rule{0pt}{9pt} & 355M                   & T+L           &   \textbf{86.11}   &  \textbf{97.42}  & 80.14 \\ \toprule

\end{tabular}
\caption{
Comparison of existing models on the FUNSD, CORD, and DocVQA datasets. T/L/I denotes the "text/layout/image" modality. Grids in \colorbox{BF}{\phantom{+}} indicate that the model uses manually-annotated semantic groups. The superscript \textsuperscript{†} indicates that the model uses text segments provided by Microsoft Read API. The superscript $^\ddagger$ indicates that the model was trained with additional QA data to achieve higher scores, it isn't directly comparable.
}
\label{table:sota}
\end{table*}

\section{Experiments}
\subsection{Pre-training Settings}

We pre-train ReLayout on the IIT-CDIP Test Collection \cite{lewis2006building}, which contains over 11 million scanned document pages. We extract words and global 1D positions, word-wise 2D bounding boxes, and text segments from document page images with an open-source OCR tool, PaddleOCR.\footnotetext{\url{https://github.com/PaddlePaddle/PaddleOCR}}

ReLayout's model architecture is almost the same as RoBERTa \cite{liu2019roberta} but with an additional 2D embedding layer. All parameters, except for the 2D embedding layer, are initialized with RoBERTa's parameters. We use AdamW optimizer \cite{loshchilov2017decoupled} with a batch size of 32 for 5 epochs. The base learning rate is set to 5e-5, with weight decay of 1e-2 and $(\beta_1, \beta_2) = (0.9, 0.999)$. The learning rate changes with a linear decay strategy. We evaluated two variants based on RoBERTa variants, i.e., ReLayout$_\text{Base}$ and ReLayout$_\text{Large}$. The former has 12 layers with 16 heads; the latent dimensionality is 768. The latter has 24 layers with 16 heads where the latent dimensionality is 1024.

As for the hyper-parameters, $P_\text{MLM}=20$\% and $P_\text{1-LOP}=30$\%. For $L_\text{2-TSC}$, the thresholds are $\theta_\text{dis}=120$ and $\theta_\text{sim}=0.9$. The coefficients for balancing the objectives are $\alpha = 0.5$ and $\gamma = 0.5$.

\subsection{Fine-tuning Settings}
\noindent\textbf{FUNSD and CORD.} FUNSD \cite{jaume2019funsd} and CORD \cite{park2019cord} are used for semantic entity classification tasks in complex forms and receipt documents, aiming to classify words into a set of predefined semantic entities. The FUNSD dataset contains 199 documents with annotations for 9,707 semantic entities, which are among ``question,'' ``answer,'' ``header,'' and ``other''. The training and test splits contain 149 and 50 samples, respectively. CORD is a dataset for information extraction in receipts with 30 semantic labels in 4 categories. It contains 1,000 receipts, 800 for training, 100 for validation, and 100 for test. For these two datasets, we use BIO tags \cite{xu2020layoutlm} and formalize semantic entity classification as a sequential labeling task. 

We fine-tune ReLayout for 1,000 steps with learning rate of 4.5e-5 and batch size of 64 for FUNSD, while learning rate of 7e-5 and batch size of 32 for CORD. Similarly to the existing methods, we use officially-provided OCR annotations (including words, word-wise bounding boxes, and global 1D positions) on the training set and report word-level F1 scores on the test set. For models that use manually annotated semantic groups, which are used to set segment-wise bounding boxes as the model's 2D position input (shaded rows in Table~\ref{table:sota}), we also report their performance when semantic groups are replaced by text segments by Microsoft Read API (MSR)\footnote{\url{https://learn.microsoft.com/en-us/azure/ai-services/computer-vision/concept-ocr}}. 

% Note that this is not strictly a ReVrDU task, as other data have not been replaced. 

\noindent\textbf{DocVQA.}
Visual question answering on document images requires a model to take a document image (if need), OCR annotations, and a question as input and output an answer. The DocVQA dataset \cite{mathew2021docvqa} offers 10,194/1,286/1,287 images and 39,463/5,349/5,188 questions in training/validation/test splits, respectively. We formalize this task as an extractive QA problem, wherein a model predicts the start and end positions with binary classifiers. We fine-tune models on the training set and report ANLS (average normalized Levenshtein similarity), a commonly-used edit distance-based metric, on the test set. Unfortunately, the OCR annotations provided in this dataset are of low quality. We thus use the MSR to extract words, word-wise bounding boxes, and global 1D positions. For models that use segment-wise bounding boxes, we employ text segments' bounding boxes by MSR (marked with \textsuperscript{†} in Table~\ref{table:sota}). We fine-tune all models for 40 epochs with learning rate of 2e-5 and batch size of 32.

Besides the experiments above, the scores of all other models come from previous papers \cite{huang2022layoutlmv3, tu2023layoutmask}.

\subsection{Results}
As shown in Table~\ref{table:sota}, when officially-providedOCR annotations are used (and so the task is VrDU), ReLayout surpasses all models that do not use manually annotated semantic groups. The models that use manually annotated semantic groups (referred to as segment-dependent models, shaded rows in Table \ref{table:sota}) yielded higher scores. Their performance significantly drop if semantic groups are replaced with text segments provided by a commercial OCR tool   (e.g., 
the performance of LayoutLMv3$_\mathrm{Base}$ and LayoutMASK$_\mathrm{Base}$ drop by $-8.68$ and $-18.94$ respectively on FUNSD.). This implies that segment-dependent models heavily rely on manually annotated semantic groups to capture semantic structures in documents, which may contradict the initial purpose of automated document understanding. Also, the results reinforce the necessity  to reexamine the choice of semantic grouping in the VrDU tasks, which is previously proven to be a shortcut \cite{li2021structurallm}.
%\footnotetext{In Table~\ref{table:sota}, the models shaded in gray use segment-wise bounding boxes, which can be obtained either through semantic groups or text segments.}

For the DocVQA dataset, it is fair to compare ReLayout with LayoutLMv3 and LayoutMASK in the ReVrDU setting (i.e., without manually annotated semantic grouping, marked with \textsuperscript{†} in the DocVQA column of Table \ref{table:sota}), and ReLayout outperforms them. Yet, it consistently falls behind LayoutLMv2. We believe this gap primarily stems from the absence of the visual modality, as LayoutLMv2 additionally leverages a visual encoder. %We are more interested in the comparison on the ReVrDU task, which will be discussed next.

\begin{table}[t]
\small
\renewcommand{\arraystretch}{1.5} % 增加行高
\begin{tabular}{
  >{\raggedright\arraybackslash}p{0.10\columnwidth}
  >{\centering\arraybackslash}p{0.15\columnwidth}
  >{\centering\arraybackslash}p{0.14\columnwidth}
  >{\centering\arraybackslash}p{0.14\columnwidth}
  >{\centering\arraybackslash}p{0.22\columnwidth}
}
\Xhline{1pt}  
 & \textbf{ReLayout} & \textbf{LMv2} & \cellcolor{gray!25}\textbf{LMv3} & \cellcolor{gray!25}\textbf{LayoutMASK}   \\ 
\Xhline{1pt}  
FUNSD & 84.64 & 82.76 & 90.29 & \textbf{92.91}   \\ 
\hdashline
\hspace{3mm}-M & \textbf{83.13} & 80.13 & 81.46 & 73.61   \\
\hspace{3mm}-P & \textbf{82.87} & 78.69 & 80.27 & 71.15  \\
\Xhline{1pt} 
CORD & 96.82 & 94.95 & 96.56 & \textbf{96.99}  \\
\hdashline
\hspace{3mm}-M & \textbf{96.23} & 93.78 & 94.37 & 82.33  \\
\hspace{3mm}-P & \textbf{95.92} & 92.24 & 93.87 & 81.47  \\
\Xhline{1pt} 
DocVQA & - & 78.08 & \textbf{78.76} & -  \\
\hdashline
\hspace{3mm}-M & 76.02 & \textbf{76.33} & 74.56 & 70.79  \\
\hspace{3mm}-P & \textbf{64.25} & 63.73 & 63.26 & 60.81  \\
\hspace{3mm}-O & \textbf{74.19} & 72.56 & 71.17 & 69.96  \\
\Xhline{1pt}  
\end{tabular}
\caption{\label{table:real-world-on-tasks}
Comparison in the ReVrDU setting with different OCR parsing results. M and P respectively indicate that the OCR parsing results are from MSR and PaddleOCR.  O in DocVQA is the officially provided OCR parsing results by an OCR tool (not manually annotated ones).
}
\end{table}

\subsection{Comparison with Different OCR Tools}
To evaluate how differences in real-world OCR tools affect the ReVrDU performance, we used both open-source and commercial OCR tools to extract OCR parsing results to create revised datasets. We assess ReLayout, LayoutLMv2 (LMv2), LayoutLMv3 (LMv3), and LayoutMASK (all of them are the base variant) over multiple OCR parsing results on the FUNSD, CORD, and DocVQA dataset, where the models are fine-tuned on the revised training sets and evaluated on the revised test sets. Table~\ref{table:real-world-on-tasks} summarizes the scores.

On the FUNSD-M and FUNSD-P datasets, segment-dependent models, i.e., LayoutLMv3 and LayoutMASK, show a significant performance drop, especially LayoutMASK. As we discussed in the previous section, manually annotated semantic groups offer a strong cue to capture the semantics in the document as the text within each group is complete (as shown in Figure \ref{fig:example}(top)), casting the word-level semantic entity classification into segment-level classification. When text segments provided by OCR tools are used, the models struggle to understand the semantic structure spanned over multiple text segments, leading to classification failures. The models that do not rely on manually annotated semantic groups, like ReLayout and LayoutLMv2, show only a slight performance drop. The smaller performance decline of ReLayout demonstrates its robustness against imperfect layout information provided by various OCR tools.

\begin{figure}[t]
\centering
\includegraphics[width=0.48\textwidth]{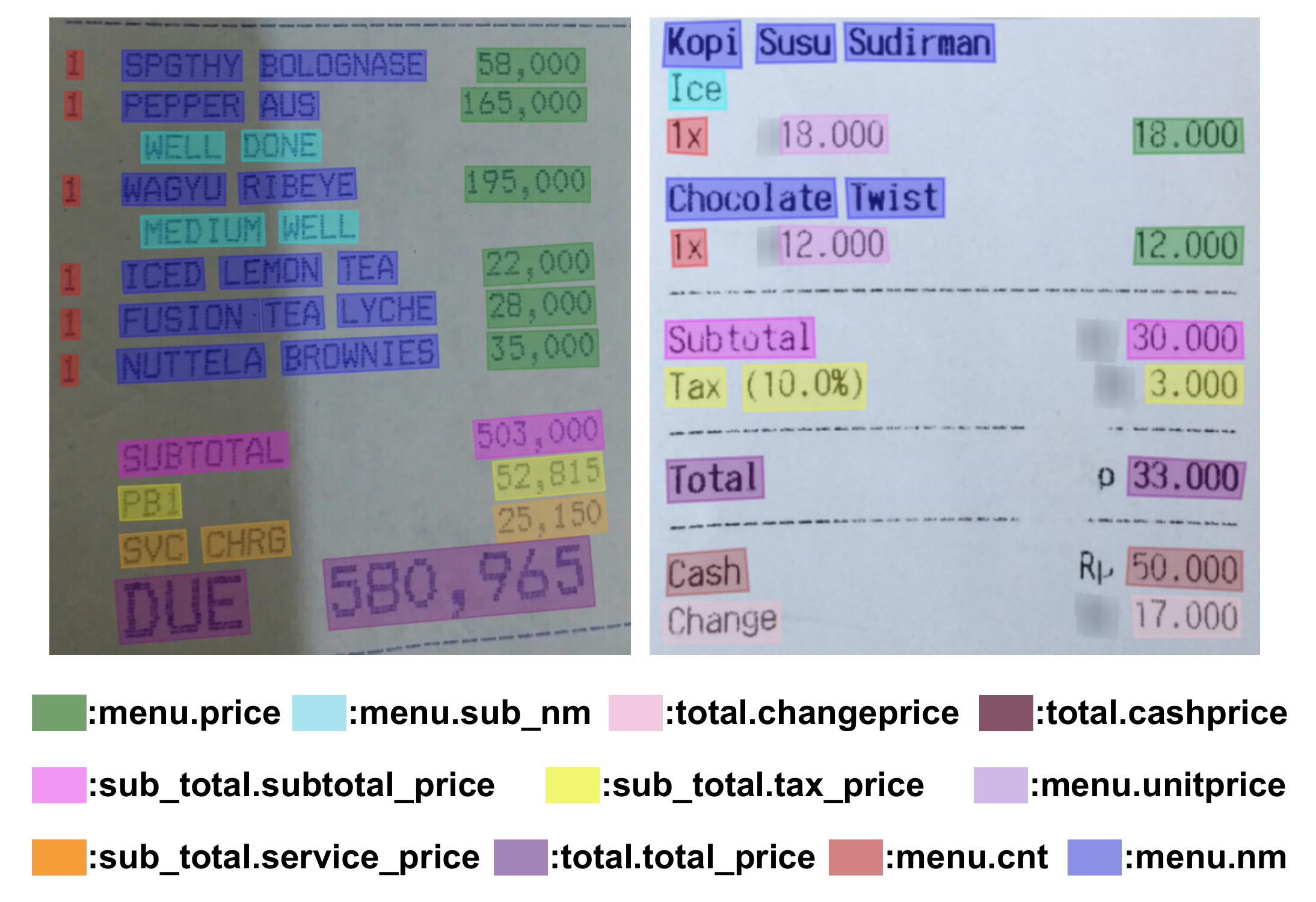}
\caption{Two examples document images from CORD.}
\label{fig:cord-sample}
\end{figure}

\begin{table*}[t]
\centering
\small
\renewcommand{\arraystretch}{1.3}
\begin{tabular}{c|ccc|cccccc}
\hline

\multirow{2}{*}{\textbf{\#}} & \multicolumn{3}{c|}{\textbf{Pre-training Setting}}        & \multicolumn{6}{c}{\textbf{Datasets}}                                \\ \cline{2-10} 

                             & \textbf{MLM} & \textbf{1-LOP} & \textbf{2-TSC} & \textbf{FUNSD} & \textbf{CORD} & \textbf{DocVQA} & \textbf{FUNSD-P} & \textbf{CORD-P} & \textbf{DocVQA-P} \\ \hline
                             
1                            &       $\surd$       &              &                         & 81.55$\pm$0.06       & 95.45$\pm$0.03    & 73.85$\pm$0.12     & 79.96$\pm$0.06  & 94.65$\pm$0.02 & 62.97$\pm$0.10          \\

2                            &         $\surd$      &        $\surd$       &              &            83.84$\pm$0.18      & 96.43$\pm$0.15    & 74.30$\pm$0.07     & 82.94$\pm$0.08  & 95.49$\pm$0.13 & 63.17$\pm$0.04            \\

3                            &          $\surd$     &           &     $\surd$          & 81.73$\pm$0.05       & 95.82$\pm$0.07    & 74.07$\pm$0.05     & 81.52$\pm$0.03  & 95.68$\pm$0.10 & 62.44$\pm$0.17          \\

4                            &       $\surd$        &          $\surd$     &     $\surd$          & 
\textbf{84.37$\pm$0.12}      & \textbf{96.64$\pm$0.12}    & \textbf{74.82$\pm$0.05}     & \textbf{83.15$\pm$0.05}  & \textbf{95.91$\pm$0.06} & \textbf{63.58$\pm$0.08}            \\ \hline
\end{tabular}
\caption{\label{table:ablation-on-tasks}
Ablation experiments of different pre-training methods.
}
\label{fig:ablation}
\end{table*}

\begin{figure}[t]
\vspace{7px}
\centering

\begin{subfigure}{0.49\textwidth}
    \centering
    \includegraphics[width=\textwidth]{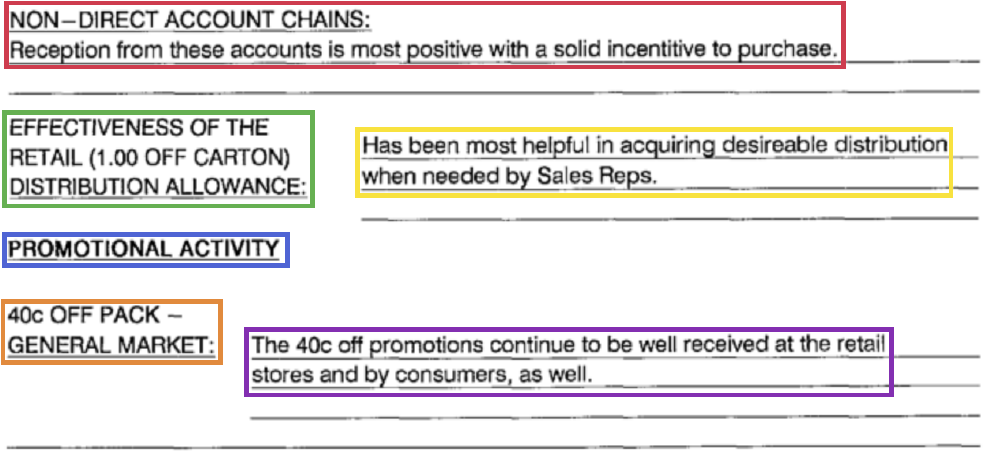}
    \label{subfig:group-text-segments-image}
\end{subfigure}

\vspace{-7px}

\begin{minipage}{0.48\textwidth}
    \centering
    \begin{subfigure}{0.48\textwidth}
        \centering
        \includegraphics[width=\textwidth]{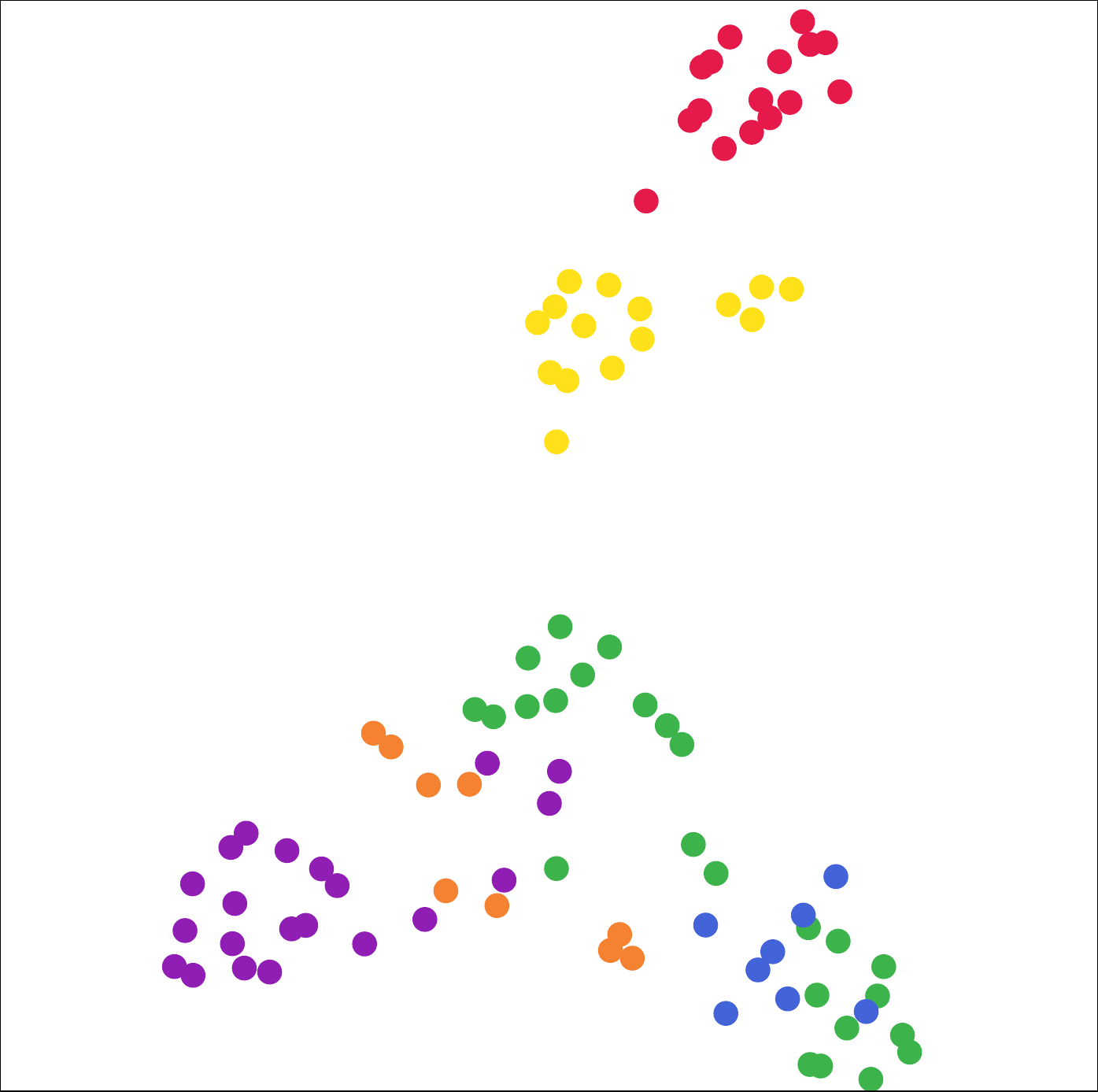}
        \caption{MLM and 1-LOP}
        \label{subfig:group-text-segments-a}
    \end{subfigure}%
    \hfill % 确保两个子图之间分布均匀
    \begin{subfigure}{0.48\textwidth}
        \centering
        \includegraphics[width=\textwidth]{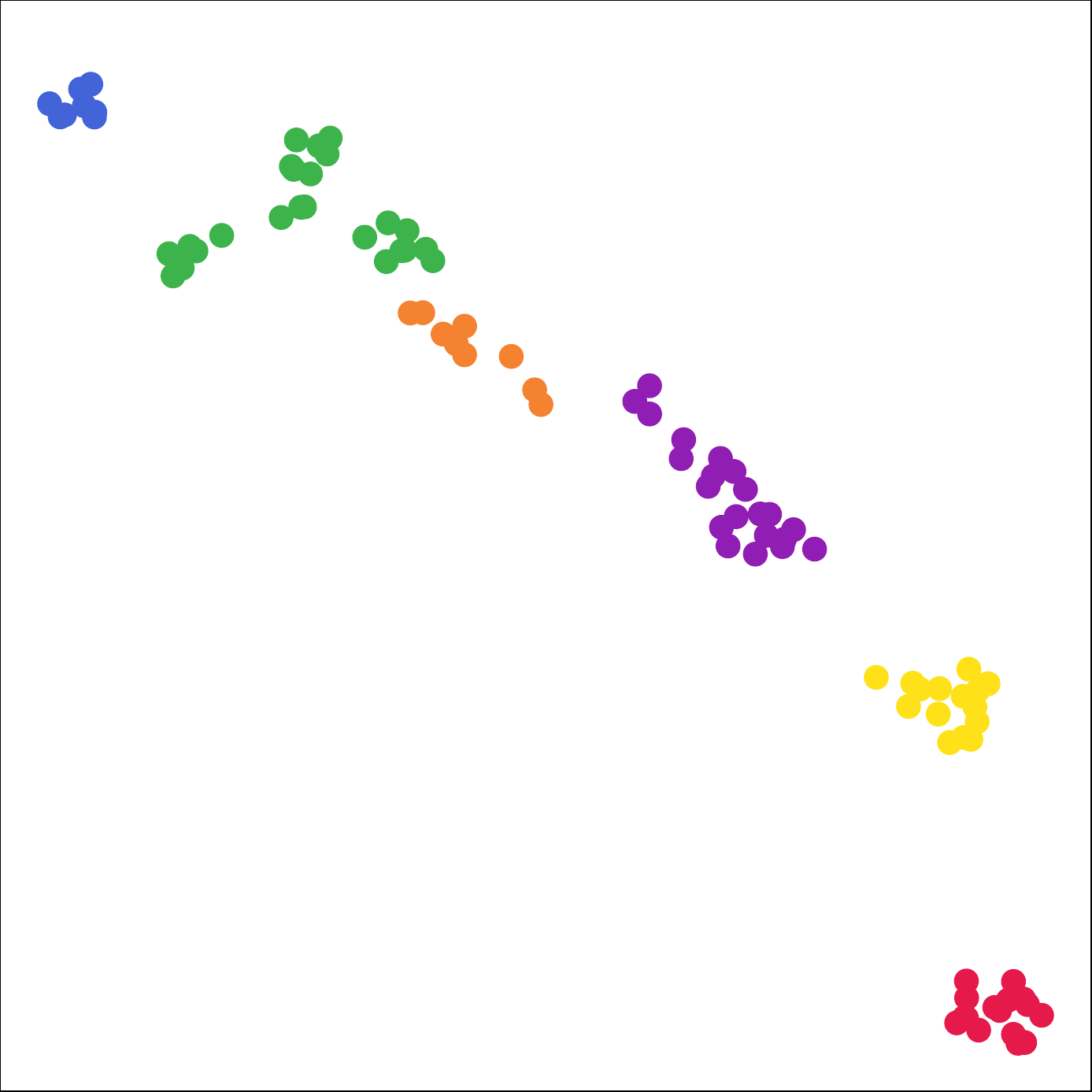}
        \caption{MLM, 1-LOP and 2-TSC}
        \label{subfig:group-text-segments-b}
    \end{subfigure}
\end{minipage}

\caption{Visualization of pre-trained representations.}
\label{fig:cluster}
\vspace{-5mm}
\end{figure}

On the CORD-M and CORD-P datasets, the four models respectively show performance declines. ReLayout still maintains the best robustness with real-world layout input. LayoutLMv3 shows an acceptable performance drop, while LayoutMASK still experiences a significant decline. The visual modality may serve as a beneficial complement when semantic grouping is inaccurate. Overall, the performance declines on CORD are smaller compared to FUNSD, possibly because the semantic ststructures receipts are basically complete in a single line as shown in Figure \ref{fig:cord-sample}, which is much simpler than those in the FUNSD.

For DocVQA, the scores in the first row of the table originate from the original reports in the LayoutLMv2 and LayoutLMv3 papers \cite{xu2020layoutlmv2, huang2022layoutlmv3}, while LayoutMASK was not evaluated on this dataset. We evaluate the models on DocVQA in the ReVrDU setting with three types of automatically acquired OCR parsing results. ReLayout shows the best performance on the revised datasets (the -P and -O variants) with poorer OCR quality, but when using the commercial MSR, ReLayout performs slightly lower than LayoutLMv2. This still demonstrates ReLayout's robustness in extracting semantically meaningful structural information even in complex documents and potentially erroneous OCR. We acknowledge that integrating the visual modality is effective in enhancing performance on theReVrDU task.

\section{Ablation Study}
We ablate newly added pre-training losses which learn comprehensively layout information in the VrDU and ReVrDU setting.  

\noindent \textbf{Quantitative analysis.} Table \ref{fig:ablation} shows the performance scores for all possible combinations of losses (the masked language model loss cannot be removed as it is the basis for pre-training). The use of 1-LOP significantly enhances model performance, particularly on the FUNSD dataset. This is because forms contain densely packed local text structures, and using 1-LOP not only helps the model enhance the comprehension of the text flow but also aids in capturing cross-segment relationships. Comparison between the first and third rows shows that 2-TSC can bring a certain, though limited, performance improvement. The combination of 2-TSC and 1-LOP improves the performance  by a larger margin. We can guess that 2-TSC hardly stand by itself as it mainly relies on the local layout information of tokens when determining relevant local text segments to bring closer. The 1-LOP loss may give ideas about the local layout information, ending up with better representations that helps 2-TSC. This is also why we only add the 2-TSC loss in the final epoch.

\noindent \textbf{Qualitative analysis.} To evaluate whether the 2-TSC pre-training task can effectively learn semantic groups, we input words, global 1D positions, and word-wise 2D bounding boxes from official annotations of a FUNSD form into the MLM and 1-LOP pre-training models with/without the 2-TSC loss. For the document shown in Figure \ref{fig:cluster}(top), we visualize the respective models' token representations using UMAP \cite{mcinnes2018umap}. The document has six (manually annotated) semantic groups, identified by boxes in different colors. Figure \ref{subfig:group-text-segments-a} displays the representations without 2-TSC, while Figure \ref{subfig:group-text-segments-b} shows those with 2-TSC (Representations learned solely from MLM pre-training can be seen in the Appendix~\ref{sec:appendix}.). The representations with MLM and 1-LOP form reasonable clusters, though the green, blue, orange, and purple semantic groups overlap to some extent. This may be because the 1-LOP loss introduces strong local layout information into the representations (as we also visualize the representations using only the MLM loss under the same input, which showed a very chaotic distribution, but due to space limitations, we did not display it). The 2-TSC loss leads to more clear-cut clusters, without using manually annotated semantic groups. This difference does not directly explain the better performance of the model with 2-TSC, but it still demonstrates that 2-TSC can be a reasonable proxy of manually annotated semantic grouping. 

\section{Conclusion}

This paper introduces the ReVrDU task that align more with real-world scenarios compared to the original VrDU tasks, shedding light on the problem of using manually annotated semantic grouping for document understanding. Our experimental results showed that the existing models worsen performance scores when accurate semantic groups are unavailable. We also propose pre-training losses, 1-LOP and 2-TCS, to aid the lack of semantic grouping, showing superior performance compared to the existing models but the reliance on semantic grouping removed. We believe the ReVrDU task brings a new dimension of challenge into document understanding and contributes its progress.

\section{Limitation and Future Work}
\noindent \textbf{Introducing the visual modality:} Incorporating information from the visual modality through a visual encoder is a common approach to enhance document understanding models. This requires considering the interactions between the three modalities: text, layout, and image. Future work will explore compatible ways to introduce visual modality information, further improving the performance of document understanding models.

\noindent \textbf{Dependency on OCR tools:} Most state-of-the-art pre-trained document understanding models rely on OCR annotations, and our model is no exception. However, this two-stage data processing approach means that the performance of OCR can significantly affect the subsequent model’s results. Therefore, exploring effective OCR-free models is an important direction to reduce accumulated errors, speed up processing, and lower computational costs.

\section{Ethics Statement}
After careful consideration, we believe that our paper does not introduce additional ethical concerns. We declare that our work complies with the \href{https://www.aclweb.org/portal/content/acl-code-ethics}{ACL Ethics Policy}.

\section{Acknowledgments}
This work was supported by World Premier International Research Center Initiative (WPI), MEXT, Japan. This work is also supported by JST ACT-X Grant Number JPMJAX24C8 and JSPS KAKENHI No. 24K20795.

\bibliography{custom}

\appendix

\section{Appendix}
\label{sec:appendix}
\subsection{Visualization of Representations}
\begin{figure*}[t] % 使用 figure* 环境来确保图形跨越两列
    \centering
    \begin{subfigure}{0.32\textwidth}
        \includegraphics[width=\linewidth]{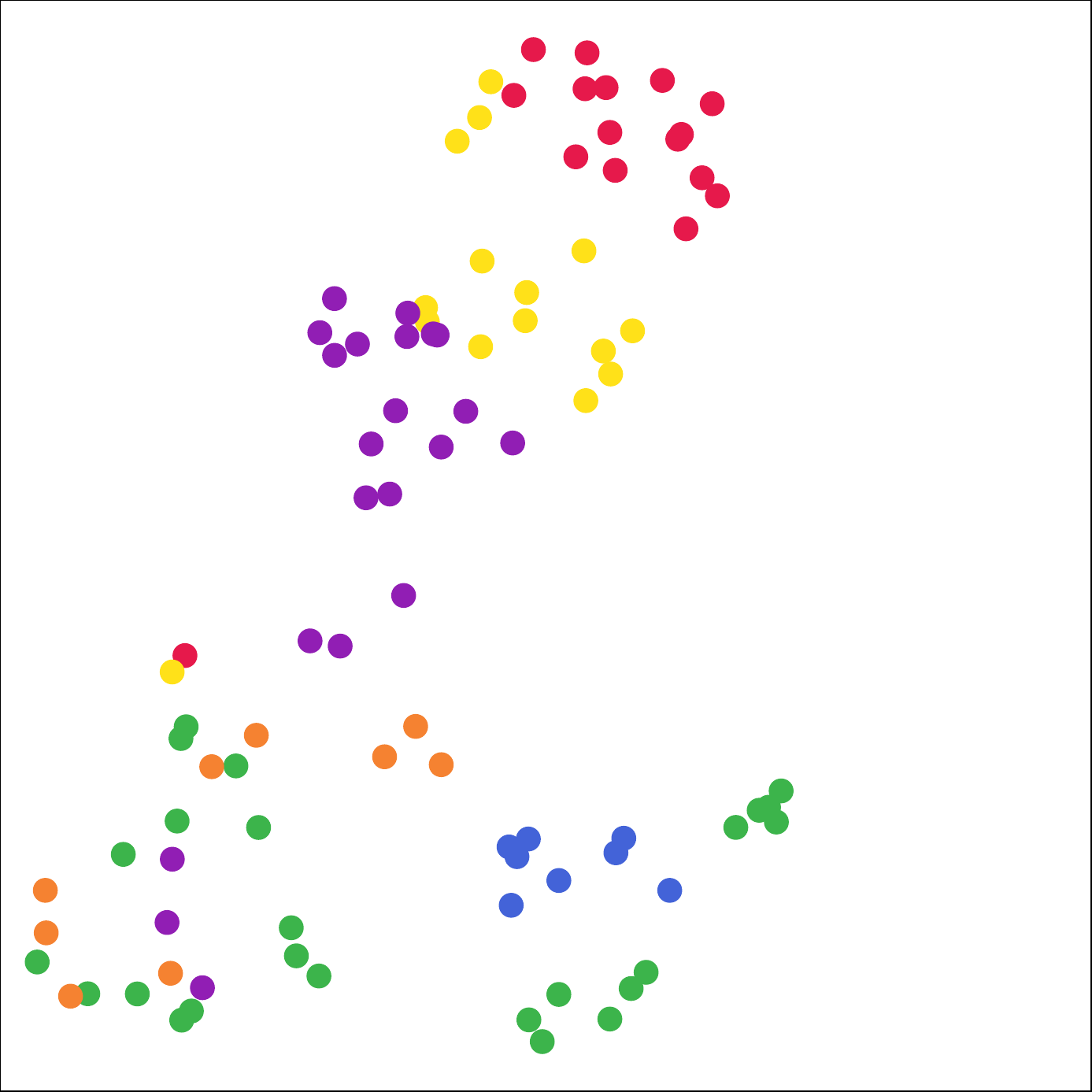}
        \caption{MLM}
    \end{subfigure}
    \hfill
    \begin{subfigure}{0.32\textwidth}
        \includegraphics[width=\linewidth]{figures/rayout_1-32.pdf}
        \caption{MLM and 1-LOP}
    \end{subfigure}
    \hfill
    \begin{subfigure}{0.32\textwidth}
        \includegraphics[width=\linewidth]{figures/rayout_12-32.pdf}
        \caption{MLM, 1-LOP and 2-TSC}
    \end{subfigure}
    \caption{Visualization of representations learned under different pre-training tasks.}
    \label{fig:complete-representation}
\end{figure*}

In Table~\ref{fig:complete-representation}, we visualize the representations learned from different pre-training task combinations.

\subsection{OCR Annotation Visualization}
Figures 6-8 display OCR annotation visualizations on the FUNSD datasets, categorizing bounding boxes into word-wise and segment-wise types, and annotations also are classified into three types: official manual, MSR, and PaddleOCR annotations. In Figure~\ref{docvqa-sample}, the images, questions, and answers from the DocVQA dataset are visualized."

%  funsd
\begin{figure*}[t] % 使用 figure* 环境来确保图形跨越两列
    \centering
    \begin{subfigure}{0.96\textwidth}
        \includegraphics[width=\linewidth]{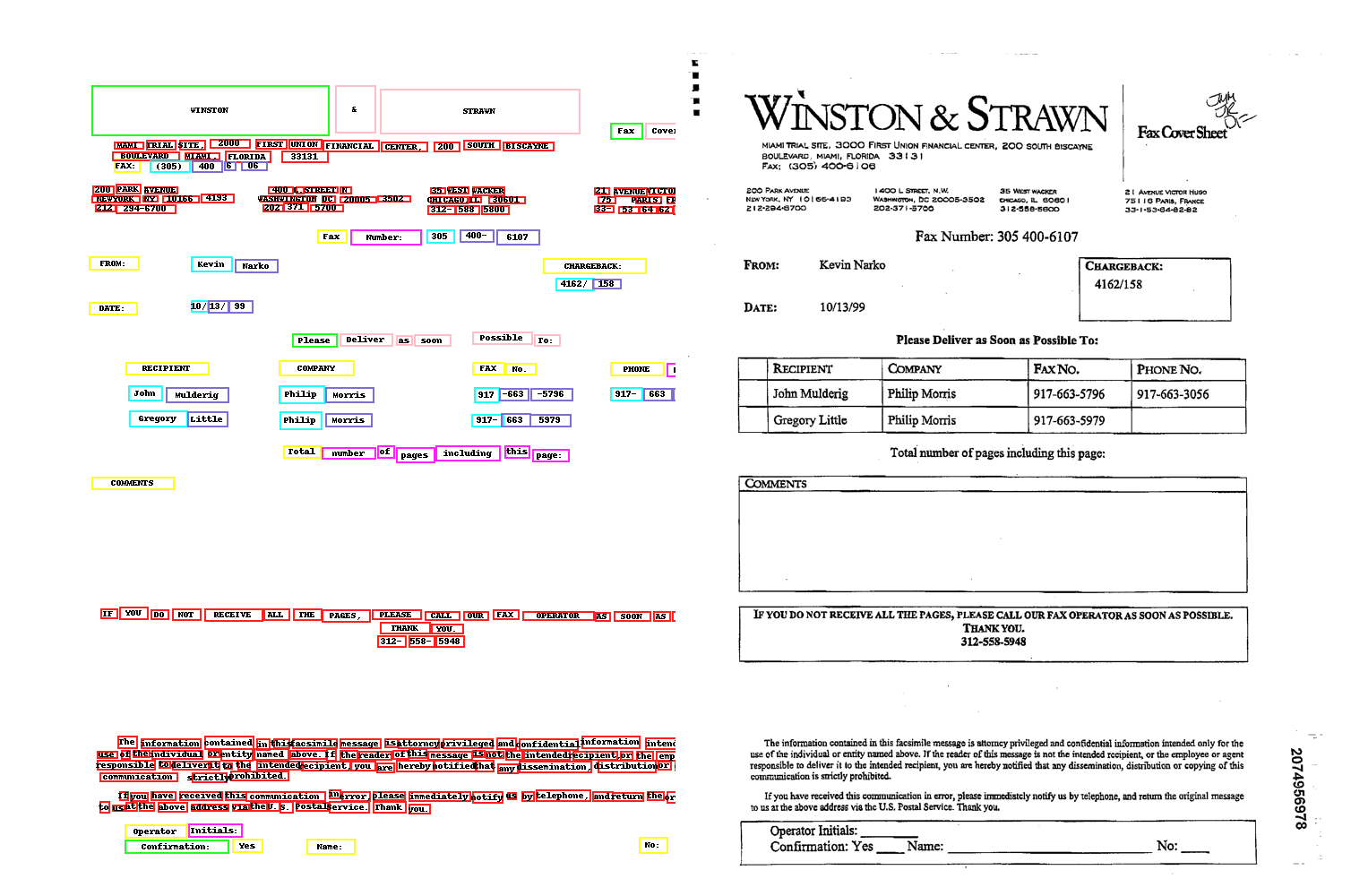}
        \caption{FUNSD-word-official-sample}
    \end{subfigure}
    \begin{subfigure}{0.96\textwidth}
        \includegraphics[width=\linewidth]{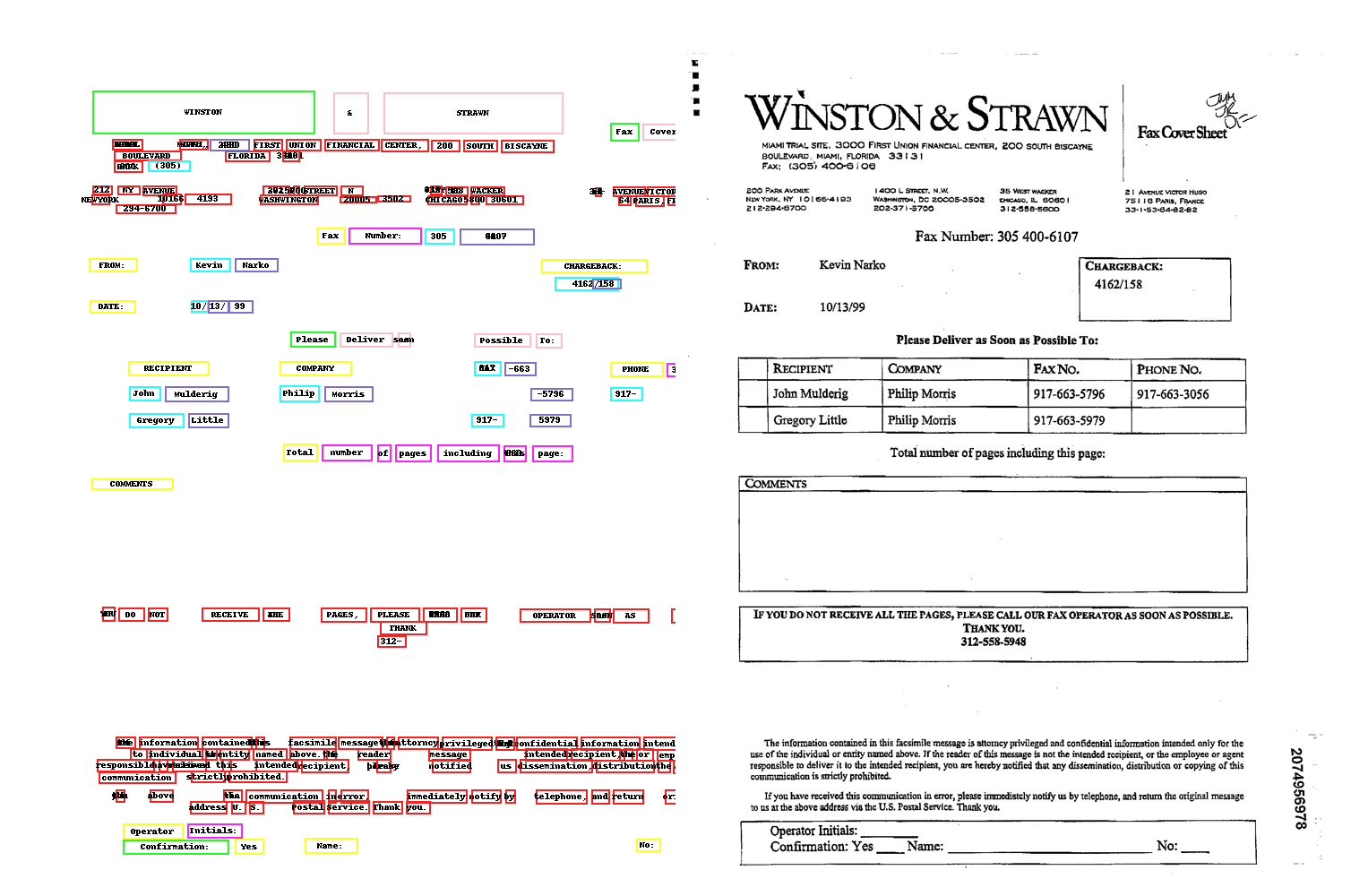}
        \caption{FUNSD-word-MSR-sample}
    \end{subfigure}
    \caption{FUNSD samples}
\end{figure*}

\begin{figure*}[t] % 使用 figure* 环境来确保图形跨越两列
    \centering
    \begin{subfigure}{0.96\textwidth}
        \includegraphics[width=\linewidth]{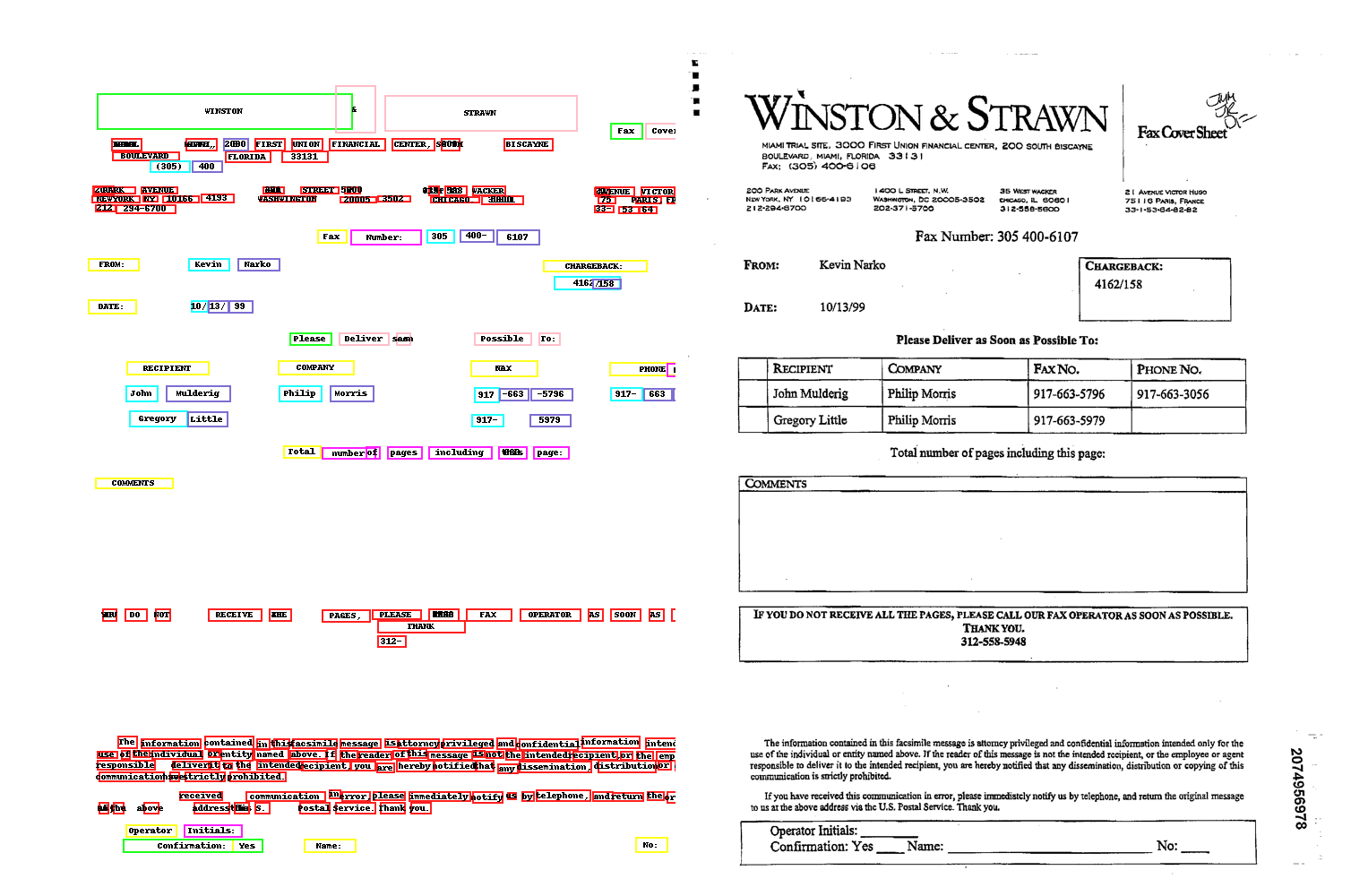}
        \caption{FUNSD-word-PPOCR-sample}
    \end{subfigure}
    \begin{subfigure}{0.96\textwidth}
        \includegraphics[width=\linewidth]{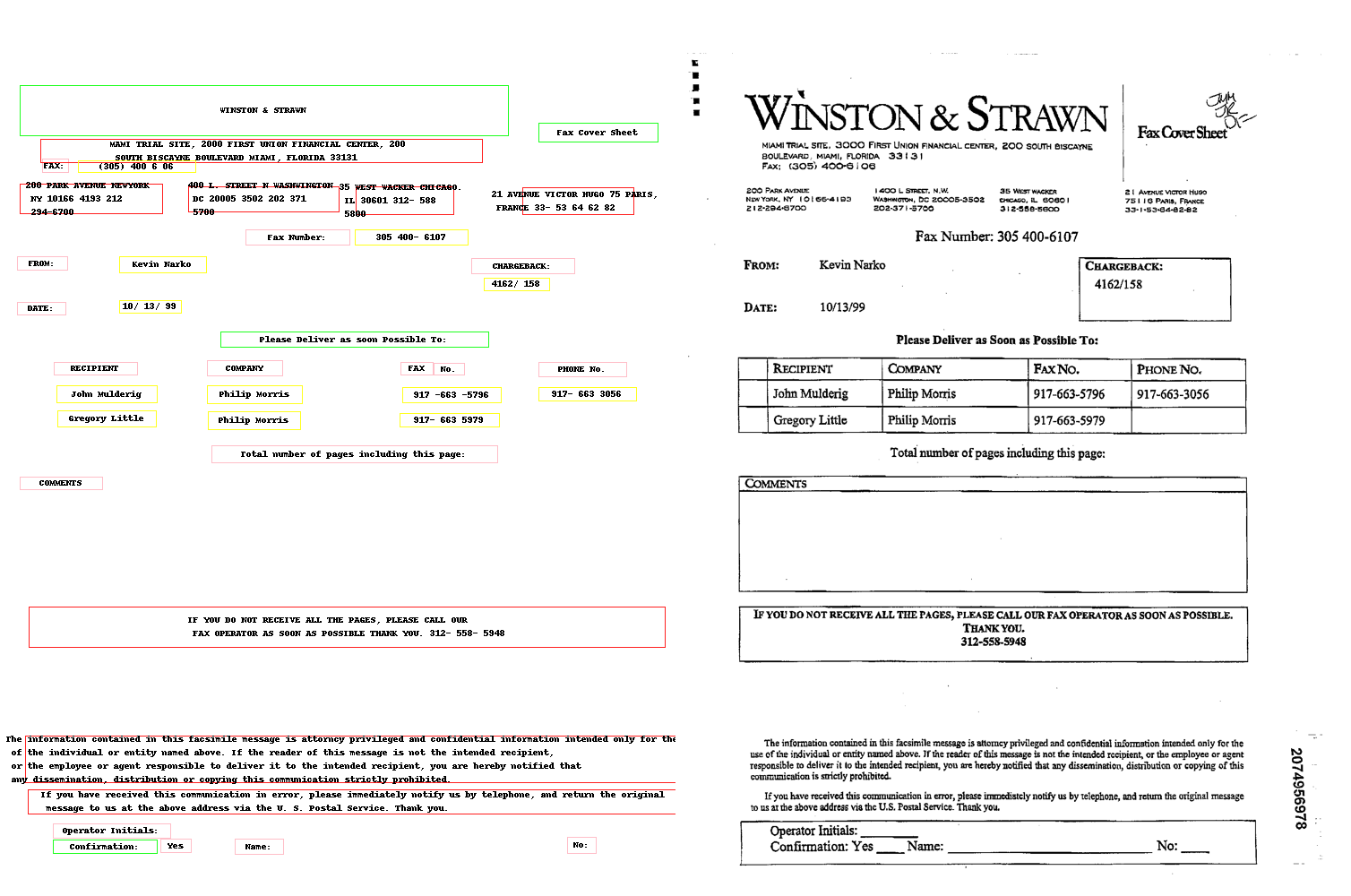}
        \caption{FUNSD-segment-official-sample}
    \end{subfigure}
    \caption{FUNSD samples}
\end{figure*}

\begin{figure*}[t] % 使用 figure* 环境来确保图形跨越两列
    \centering
    \begin{subfigure}{0.96\textwidth}
        \includegraphics[width=\linewidth]{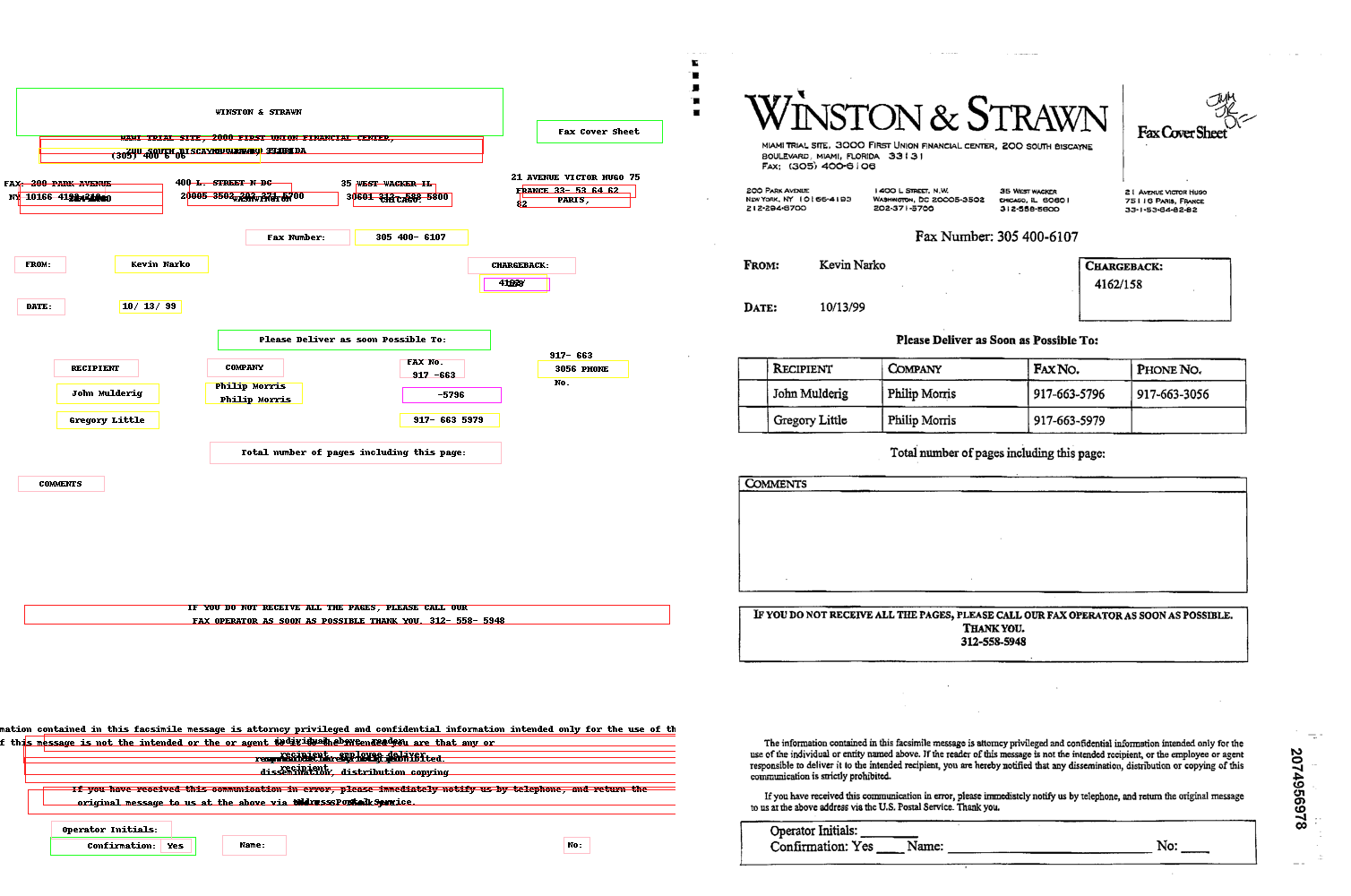}
        \caption{FUNSD-segment-MSR-sample}
    \end{subfigure}
    \begin{subfigure}{0.96\textwidth}
        \includegraphics[width=\linewidth]{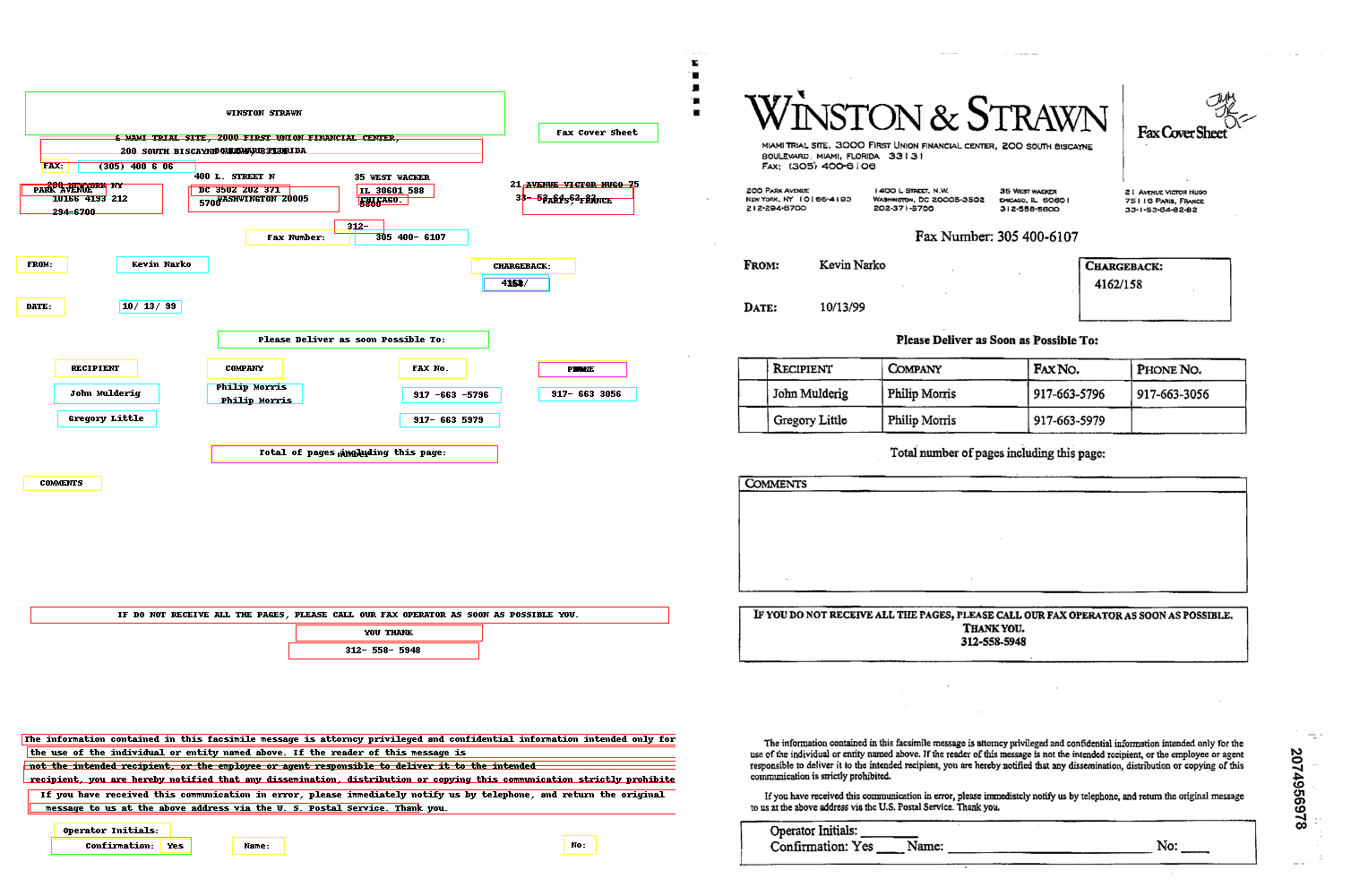}
        \caption{FUNSD-segment-PPOCR-sample}
    \end{subfigure}
    \caption{FUNSD samples}
\end{figure*}

% CORD
% \begin{figure*}[t] % 使用 figure* 环境来确保图形跨越两列
%     \centering
%     \begin{subfigure}{0.96\textwidth}
%         \includegraphics[width=\linewidth]{appendix_images/cord-word-official-sample.png}
%         \caption{CORD-word-official-sample}
%     \end{subfigure}
%     \begin{subfigure}{0.96\textwidth}
%         \includegraphics[width=\linewidth]{appendix_images/cord-word-MSR-sample.png}
%         \caption{CORD-word-MSR-sample}
%     \end{subfigure}
%     \caption{CORD samples}
% \end{figure*}

% \begin{figure*}[t] % 使用 figure* 环境来确保图形跨越两列
%     \centering
%     \begin{subfigure}{0.96\textwidth}
%         \includegraphics[width=\linewidth]{appendix_images/cord-word-PPOCR-sample.png}
%         \caption{CORD-word-PPOCR-sample}
%     \end{subfigure}
%     \begin{subfigure}{0.96\textwidth}
%         \includegraphics[width=\linewidth]{appendix_images/cord-segment-official-sample.png}
%         \caption{CORD-segment-official-sample}
%     \end{subfigure}
%     \caption{CORD samples}
% \end{figure*}

% \begin{figure*}[t] % 使用 figure* 环境来确保图形跨越两列
%     \centering
%     \begin{subfigure}{0.96\textwidth}
%         \includegraphics[width=\linewidth]{appendix_images/cord-segment-MSR-sample.png}
%         \caption{CORD-segment-MSR-sample}
%     \end{subfigure}
%     \begin{subfigure}{0.96\textwidth}
%         \includegraphics[width=\linewidth]{appendix_images/cord-segment-PPOCR-sample.png}
%         \caption{CORD-segment-PPOCR-sample}
%     \end{subfigure}
%     \caption{CORD samples}
% \end{figure*}

\begin{figure*}[t] % 使用 figure* 环境来确保图形跨越两列
    \centering
    \includegraphics[width=\linewidth]{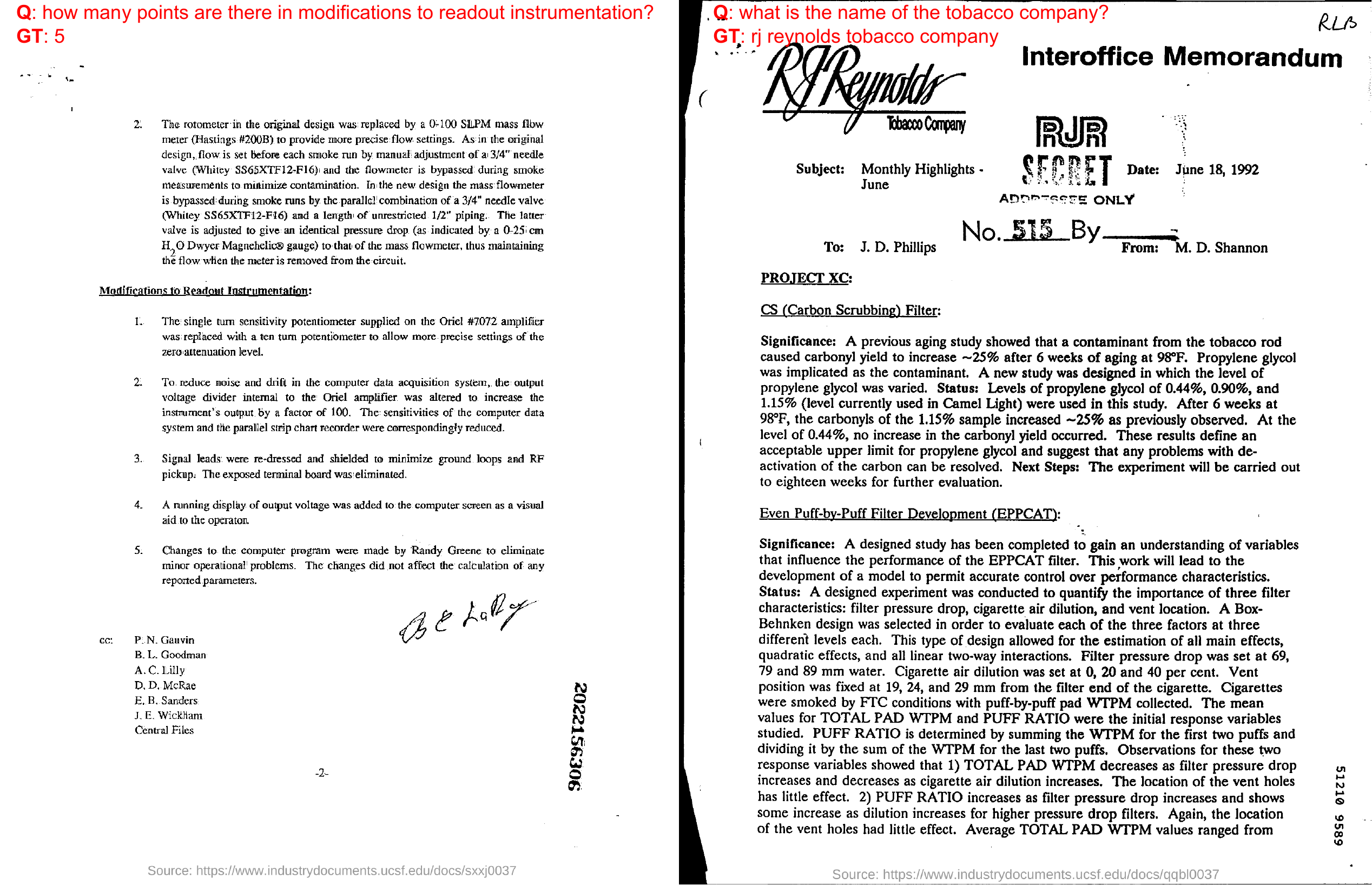}
    \caption{DocVQA samples}
    \label{docvqa-sample}
\end{figure*}

% \subsection{Representations of different pre-training tasks}

\subsection{BIO Tags}
The BIO tagging scheme is a method used for marking up text in sequence labeling tasks, commonly applied in Named Entity Recognition (NER) and other forms of linguistic annotation. In this scheme, each token of the text is tagged with one of three prefixes: "B-" (Beginning), "I-" (Inside), and "O" (Outside). The "B-" prefix indicates the beginning of an entity, "I-" marks the continuation of an entity, and "O" denotes a token that does not belong to any entity. This method helps in clearly differentiating the boundaries of entities within the text, making it easier for models to recognize and categorize text segments accurately. For example, in the entity "New York," "New" would be tagged as "B-Location" and "York" as "I-Location," clearly identifying the entire phrase as a geographical entity.

Therefore, learning to use 1-LOS to learn the local reading order is beneficial for the model to solve entity classification tasks based on BIO tagging.
\end{document}